\documentclass[conference]{IEEEtran}
\IEEEoverridecommandlockouts

\usepackage{multirow} 

\usepackage{cite}
\usepackage{amsmath,amssymb,amsfonts}
\usepackage{algorithmic}
\usepackage{graphicx}
\usepackage{textcomp}
\usepackage{xcolor}
\def\BibTeX{{\rm B\kern-.05em{\sc i\kern-.025em b}\kern-.08em
    T\kern-.1667em\lower.7ex\hbox{E}\kern-.125emX}}
\begin{document}

\title{Refining Transcripts With TV Subtitles by Prompt-Based Weakly Supervised Training of ASR}
\thanks{Foundation Flanders (FWO) under grant S004923N of the SBO programme.}

\author{\IEEEauthorblockN{1\textsuperscript{st} Xinnian Zhao}
\IEEEauthorblockA{\textit{Department Electrical Engineering ESAT-PSI} \\
\textit{KU Leuven University}\\
Leuven, Belgium \\
xzhao1@esat.kuleuven.be}
\and
\IEEEauthorblockN{2\textsuperscript{nd} Hugo Van Hamme}
\IEEEauthorblockA{\textit{Department Electrical Engineering ESAT-PSI} \\
\textit{KU Leuven University}\\
Leuven, Belgium \\
hugo.vanhamme@esat.kuleuven.be}
}

\maketitle

\begin{abstract}
This study proposes a novel approach to using TV subtitles within a weakly supervised (WS) Automatic Speech Recognition (ASR) framework. Although TV subtitles are readily available, their imprecise alignment with corresponding audio limits their applicability as supervised targets for verbatim transcription. Rather than using subtitles as direct supervision signals, our method reimagines them as context-rich prompts. This design enables the model to handle discrepancies between spoken audio and subtitle text. Instead, generated pseudo transcripts become the primary targets, with subtitles acting as guiding cues for iterative refinement. To further enhance the process, we introduce a weighted attention mechanism that emphasizes relevant subtitle tokens during inference. Our experiments demonstrate significant improvements in transcription accuracy, highlighting the effectiveness of the proposed method in refining transcripts. These enhanced pseudo-labeled datasets provide high-quality foundational resources for training robust ASR systems.
\end{abstract}


\section{Introduction}
\label{sec:intro}

Recently, foundation models have emerged as a dominant force in the field of artificial intelligence, offering extensive knowledge bases for various tasks and modalities~\cite{liu2023summary}.
These large-scale pre-trained models have significantly enhanced the performance of many downstream applications by leveraging vast datasets~\cite{moor2023foundation}.
However, the datasets used in these models are not uniform across all tasks or domains, leading to discrepancies in performance~\cite{10111523,song2023comprehensive}. 
Whisper, a widely known foundation speech model, faces the same challenge~\cite{radford2023robust}. 
It is trained on a large amount of web data along with a small portion of labeled data with weak supervision. 
Imbalances in data sizes and quality pose great challenges in generalizing the Whisper model effectively to underrepresented domains and languages~\cite{pratama2024analysis,roudit2023comparison,sehar2025benchmarking}. 
Fine-tuning on labeled data from the target domain or language is a common strategy to bridge the gap\cite{jain2024exploring}.
However, in low-resource scenarios, obtaining sufficient labeled data for real-world applications remains a challenge.
To address this, we explore methods to fine-tune the pre-trained Whisper in a weakly supervised (\textbf{WS}) framework, and aim to refine the generated transcripts with TV subtitles. 

TV subtitles are a readily available resource, featuring concise and clear text specifically designed for readability~\cite{ali2019mgb}.
They are often used as weak labels for ASR training due to their lack of precise alignment with audio~\cite{singh2020large, cheng2020weakly}. However, these methods typically require subsequent fine-tuning with verbatim transcripts to meet transcription accuracy objectives.

In this study, we use subtitles as prompts for the Whisper text decoder instead of taking them as training targets. We experiment on Flemish, a Dutch dialect. The primary weak labels for training are pseudo transcripts generated by the pre-trained Whisper model. Subtitles, acting as prompts, provide additional contextual cues for transcription, facilitating the iterative refinement of transcripts.
To further enhance performance, we introduce a weighted attention (\textbf{WA}) mechanism during inference. This mechanism selectively emphasizes relevant words in subtitle prompts, aligning them more closely with the spoken audio while disregarding irrelevant words. The present work on integrating subtitle prompting (\textbf{SP}) training and WA inference yields measurable improvements in word error rate (WER), demonstrating the effectiveness of our approach in creating higher-quality datasets for weakly supervised ASR applications without requiring additional labeled data. The key contributions of this paper are:
\begin{itemize}
    \item Development of a training methodology that incorporates SP within a WS frame work, handling the misalignment between subtitles and speech.
    \item  Introduction of a WA mechanism for inference, enhancing the model’s ability to extract relevant information from subtitles.
    \item Demonstration of how this approach improves dataset quality for WS ASR, with a focus on underrepresented and low-resource domains.
\end{itemize}

\section{Related Work}
The generative text decoder in Whisper supports prefix prompting, 
which has inspired several studies exploring its potential across various tasks. \cite{peng2023prompt,li2024prompting} designed particular prompts to adapt Whisper for zero-shot tasks,
while \cite{liao2023zero,10447492,li2024whisper} fine-tuned pre-trained Whisper models using domain-specific prompts. 
Our work differs from these studies in several key aspects. First, the lack of precise verbatim transcripts necessitates training in a weakly supervised setting.
Second, subtitles are unsuitable as direct prompts for a pre-trained Whisper model, as their overlap with audio content can mislead the model, causing truncated outputs.
Finally, a WA mechanism is introduced during inference, which addresses the unique challenges of using subtitle prompts.

\section{Methods}


\subsection{SP Training}
The objective of the task is to condition transcript generation on subtitle prompts as contextual information. Any model with a generative decoder is suitable for this purpose. We select Whisper due to its robust weakly supervised training on large-scale, multilingual web-sourced data, which enhances its transferability to our spontaneous Flemish broadcast audio. 

\begin{figure}[thb]
\centering
\centerline{\includegraphics[width=6cm]{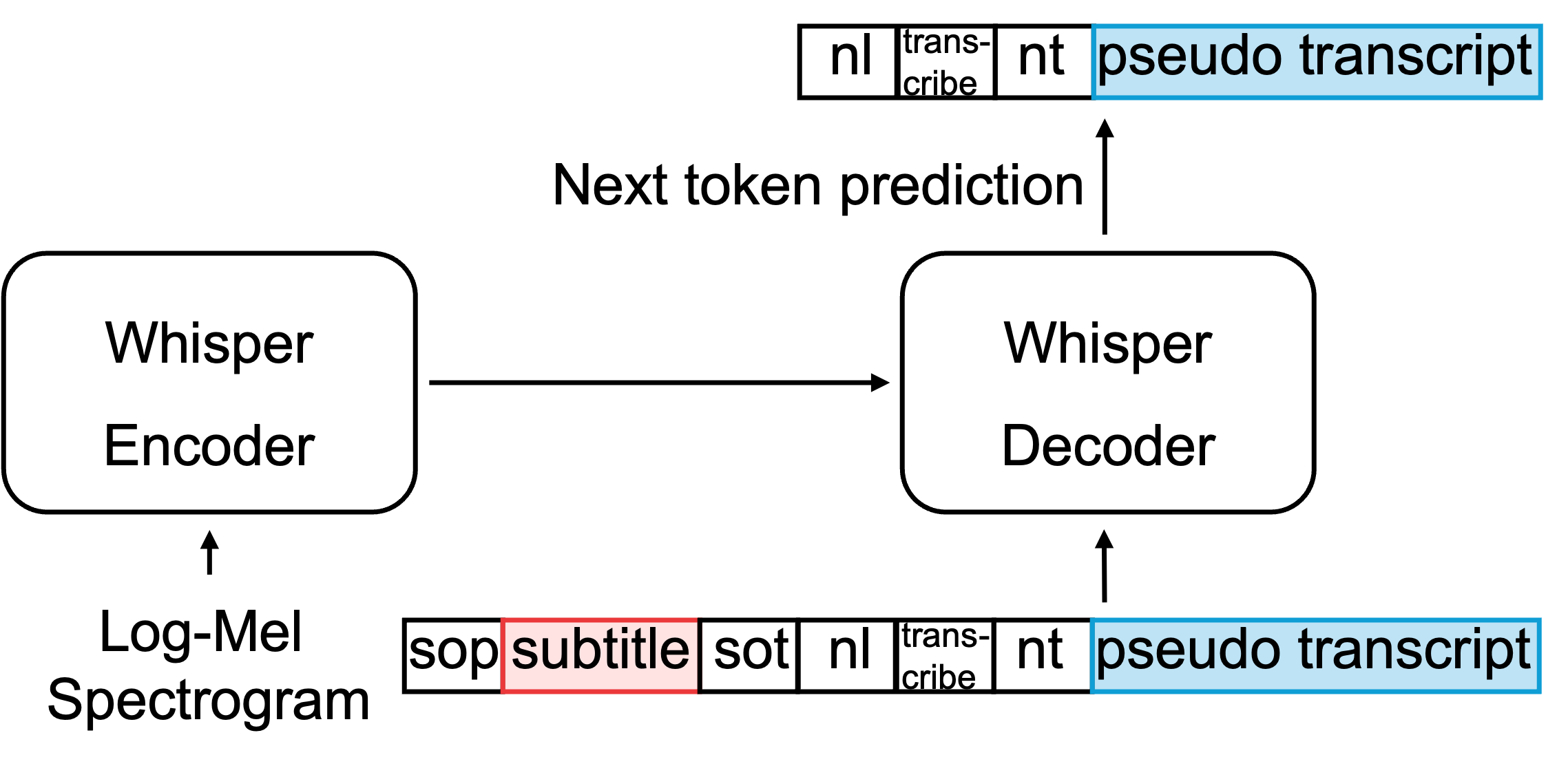}}
\caption{Framework of subtitle prompting Whisper.}
\label{fig1}
\end{figure}

Figure~\ref{fig1} illustrates the framework for Whisper SP, which integrates subtitles as context prompts and pseudo transcripts as training targets to fine-tune the model. 
To leverage transfer learning from generic Dutch, we fine-tune the \textless\textbar\verb|nl|\textbar\textgreater\quad language token using the Flemish training data.
The final input to the decoder during training is structured as 
\textless\textbar\verb|sop|\textbar\textgreater
\verb|subtitles|
\textless\textbar\verb|sot|\textbar\textgreater
\textless\textbar\verb|nl|\textbar\textgreater
\textless\textbar\verb|transcribe|\textbar\textgreater
\textless\textbar\verb|notimestamps|\textbar\textgreater.

We denote the subtitled dataset as ${X, Y_s}$, where $X$ is the audio and $Y_s$ is the paired subtitles. The pseudo transcripts $Y_{pt_0}$ are first generated for $X$ using a pre-trained Whisper model.
Whisper is known to hallucinate by repeating the same word in its output~\cite{koenecke2024careless}, especially when the audio is heavily overlapped. We remove these rare errors by filtering based on transcript length.
The filtered transcripts serve as the starting point for the iterative training process. With each iteration, the model refines the pseudo transcripts by leveraging the contextual information provided by the subtitle prompts. After training over the entire dataset once, the updated pseudo labels are denoted as $Y_{pt_1}$. This process is repeated over $t$ iterations, gradually evolving the pseudo labels from $Y_{pt_0}$ to $Y_{pt_t}$.

It should be noted that the proposed iterative training differs from self-training, where only pseudo transcriptions are used for training, requiring error control strategies to prevent error propagation \cite{amini2025self}. In our approach, subtitle prompts provide additional information, while the loss is computed on the transcript. This allows the model to progressively refine the pseudo transcripts until the information in subtitles is fully extracted.

\subsection{WA inference}
Subtitles often overlap with speech content while also containing extraneous information not reflected in the audio. The relevance of subtitle tokens to speech can be evaluated using cross-attention weights. Tokens highly relevant to speech are expected to exhibit strong and concentrated cross-attention weights to specific speech frames. In contrast, distraction tokens tend to have relatively uniform attention weights spread across all frames. Intuitively, relevant tokens are better suited to guide transcription through self-attention. Based on this principle, we aim to select relevant tokens while minimizing the influence of irrelevant ones.

To achieve this, we introduce a WA mechanism that employs the Gini coefficient \cite{dorfman1979gini} as a measure of relevance. The Gini coefficient, commonly used in economics to measure inequality in a distribution, is adapted here to quantify the distribution of cross-attention weights over input speech frames. The formula for the Gini coefficient is defined as:

\begin{equation}
g_i=\frac{\sum_{k=1}^{N}(2k-N-1) \cdot \mathcal{CA}[i,k]}{N \cdot \sum_{k=1}^{N}\mathcal{CA}[i,k]},
\label{eq1}
\end{equation}

\noindent where $\mathcal{CA}$ represents the cross-attention weight matrix, $i$ denotes a specific token in the subtitle prompt, $k$ indexes speech features of the input frames, and $N$ is the total number of speech frames. The cross-attention weights $\mathcal{CA}$ are derived from the first cross-attention layer, where higher-level text representations have not yet been computed, providing a clearer and more direct mapping from text tokens to audio inputs. At this layer, the attention map exhibits the most straightforward monotonic alignment between text and speech. 
Before computing $g_i$, the values of $\mathcal{CA}[i,k]$ are sorted cross N frames to ensure a positive value for $g_i$. The final value of $g_i$ ranges from $0$, indicating that $\mathcal{CA}$ has a uniform distribution, to $1$, signifying that $\mathcal{CA}$ exhibits a highly unequal distribution, resembling a sharp focus.

The Gini coefficient calculated over the prompt sequence in length $T_p$ is denoted as $\mathcal{G}$, which is then utilized to weight the prompt values in self-attention layers as follows:

\begin{equation}
K’= \mathcal{G} K_{p} \oplus K_{t}; \quad V’= \mathcal{G} V_{p} \oplus V_{t},
\label{eq2}
\end{equation}

\noindent Here, $K$ and $V$ represent the key and value matrices, respectively, with subscripts $p$ and $t$ denoting the prompt and transcript targets.
The matrices $K_p$ and $K_t$, with $p$ and $t$ sequences in lengths of $T_p$ and $T_t$, are concatenated along the time dimension using $\oplus$, resulting in a final sequence of length T that matches with the input length.
Thus, the dimensions of the final $K'$ and $V'$ matrices remain unchanged.
This weighting mechanism ensures that subtitle tokens most relevant to the speech input are emphasized during the attention computation.

The final self-attention output is then calculated as:
\begin{equation}
H = \text{softmax}\left(\frac{QK’^{\top}}{\sqrt{d}}\right)V’,
\label{eq3}
\end{equation}
\noindent where $Q$ represents the query derived from the prompt and the transcript, and $d$ denotes the model dimension. Through Eq.~\ref{eq3}, the model is guided to prioritize speech-relevant tokens in subtitle prompts while ignoring distractions, thereby enhancing transcription accuracy.

\section{Experiments}
\subsection{Data}
The subtitled dataset used in this study consists of 760 hours of multi-genre recordings from 16 TV programs broadcast by the Vlaamse Radio- en Televisieomroeporganisatie (VRT, Flemish Radio and Television Broadcasting Organisation) between September 2020 and November 2022. These recordings span four primary genres—news, talk shows, entertainment, and drama—covering a broad spectrum of topics, including politics, economics, education, culture, and sports. 
Most audio segments in the dataset are around 15 seconds long, with a maximum duration of 30 seconds, aligning with the input window requirements of Whisper. These segments are typically longer than the corresponding subtitle timings, which are kept short for readability. No additional pre-processing was performed on the audio; non-speech sounds, such as music, applause, and ringtones, are preserved within subtitle intervals. This simplifies the speech recognition pipeline while retaining the inherent challenges posed by spontaneous speech.


For reliable evaluation, 6 hours of speech (approximately 2,600 utterances) from each genre were manually annotated with verbatim transcripts. This annotated subset, referred to as \textit{subs-annot}, was excluded from the training set. To evaluate the suitability of the subtitles for the ASR task, the WER of the subtitles in \textit{subs-annot} was computed using the verbatim transcripts as the reference. The resulting WER of 34.3\% highlights the misalignment between subtitles and the speech content, motivating our decision to use subtitles as prompts rather than direct training targets.

\subsection{Training details}
The models were implemented using the HuggingFace Transformers library. Training was conducted on a single Nvidia H100 GPU with an effective batch size of 64, using the Adam optimizer with a learning rate of $1 \times 10^{-5}$ and 1,000 warmup steps. The models were trained iteratively across multiple epochs.

\section{Results and Discussions}
\subsection{SP Training}
Experiments were conducted to enhance transcripts using SP without requiring verbatim data. To demonstrate that SP provide contextual clues for transcript generation, we compared the WERs (\%) on the \textit{subs-annot} set across pre-trained Whisper models (medium and large variants) and models fine-tuned with or without SP. The results are presented in Table~\ref{tab2}.

\begin{table}[hbt]
\centering
\caption{Comparison of WERs (\%) on the \textit{subs-annot} set between pre-trained, no-prompting (NP) fine-tuned and SP fine-tuned Whisper medium or large models.}
\label{table}
\setlength{\tabcolsep}{5pt} 
\renewcommand{\arraystretch}{1.4} 
\begin{tabular}{c|c|c}
\noalign{\hrule height 1pt}
Model & Whisper-medium & Whisper-large-v3 \\
\noalign{\hrule height 1.0pt}
Pre-trained & 18.75 & 13.07 \\
NP Fine-tuned & 13.05 & 21.49 \\
SP Fine-tuned & \textbf{11.54} & \textbf{11.37} \\
\noalign{\hrule height 1pt}
\end{tabular}
\label{tab2}
\end{table}

The pre-trained Whisper models exhibit strong zero-shot performance on our Flemish broadcast dataset, achieving WERs below 20\% for both the medium and large models. These results indicate that the transcripts are nearly intelligible and highlight Whisper’s impressive generalization capabilities.
We did not report results for directly prompting the pre-trained Whisper models with subtitles, as this configuration consistently produced truncated transcripts or no output at all. This behavior is likely due to significant overlap between subtitle prompts and speech content. The model tends to align the subtitle prompts with the audio and avoids repeating the prefix, resulting in incomplete outputs. This observation underscores the limitations of Whisper’s text decoder in handling prompts without careful design and emphasizes the importance of tailoring prompts to prevent disruptions in transcript generation.

We transcribed all the audio in the training set using the pre-trained Whisper-large-v3 model. The resulting pseudo transcripts $Y_{pt_0}$ were used in the first iteration of fine-tuning. Not surprisingly, self-training the Whisper-large-v3 model with $Y_{pt_0}$ resulted in a significant increase in WER, as shown in the second row of Table~\ref{tab2}. This suggests error propagation, which is a well-known challenge in self-training \cite{amini2025self}. We did not apply any error control strategies, as our goal was to demonstrate the effectiveness of SP in refining errors in pseudo transcripts. Analyzing the errors in the outputs reveals that deletions account for more than 80\% of the total errors. Specifically, the outputs show that the model often fails to transcribe entire utterances or their beginnings. This issue is particularly pronounced in utterances that initially had deletions at the start in $Y_{pt_0}$, potentially caused by inaccuracies in segmenting speech. Consequently, the model struggles with determining when to begin transcribing. As a result, the large model, being highly parameterized, may inadvertently reinforce its own biases and errors during self-training, leading to degraded performance.

In contrast, fine-tuning the Whisper-medium model with $Y_{pt_0}$ led to improved performance. This is likely because the medium model had more capacity to benefit from  $Y_{pt_0}$, considering the WER gap between the pre-trained medium and large models. 
Incorporating SP into fine-tuning effectively addresses the self-training challenge in large models. The WER decreased for both the medium model and the large model compared to fine-tuning without prompts, even though the models were trained for only one iteration. This demonstrates that subtitles provide additional information to guide generation, leading to more refined transcripts as a result.

\begin{table}[hbt]
\centering
\caption{rWERs (\%)  and oWERs (\%) tested on rare words and out-of-vocabulary words from the test set.}
\label{table}
\setlength{\tabcolsep}{5pt} 
\renewcommand{\arraystretch}{1.4} 
\begin{tabular}{c|c|c|c|c}
\noalign{\hrule height 1pt}
\multirow{2}{*}{Model} & \multicolumn{2}{c|}{Whisper-medium} & \multicolumn{2}{c}{Whisper-large-v3} \\
\cline{2-5}
    &  rWER & oWER & rWER & oWER \\
\noalign{\hrule height 0.6pt}
Pre-trained & 38.42 & 77.34 & 29.74 & 71.94 \\
NP Fine-tuned & 31.04 & 74.94 & 31.90 & 72.32 \\
SP Fine-tuned & \textbf{24.63} & \textbf{70.22} & \textbf{24.23} & \textbf{69.64} \\
\noalign{\hrule height 1pt}
\end{tabular}
\label{tab3}
\end{table}

Although subtitles do not provide the exact speech content, they often include the most informative words necessary for comprehension, such as specific named entities. These words are typically rare in the training set and therefore challenging to transcribe. To gain deeper insight into the information provided by subtitles, we break down the WERs in Table~\ref{tab2} into rare word error rate (rWER) and out-of-vocabulary (OOV) word error rate (oWER), with the results presented in Table~\ref{tab3}. Here, rare words are defined as words with frequency lower than 10 in the training set, while OOV words are completely absent from the training set. There are 1,762 rare words and 1,333 OOV words out of 76,684 words in the reference transcripts of \textit{subs-annot} set.

From Table~\ref{tab3}, OOV words exhibit higher WERs than rare words overall, which aligns with intuition. SP consistently benefits the transcribing in terms of both rWER and oWER. When integrating SP to fine-tuning,
the rWER decreasing significantly from 31.04\% to 24.63\%. This is notable because fine-tuning without prompts already exposed the model to rare words, resulting in an visible improvement in rWER compared to the zero-shot results from the pre-trained model. Similar decreases are observed in oWER, suggesting that SP not only help the model recall rare words but also improve its generalization ability. The large model exhibits slightly better rWER and oWER compared to the medium model after incorporating SP. Notably, SP effectively address the challenges revealed by self-training. By introducing additional information and providing guidance for predictions, subtitles help the model focus on refining its transcription output. Consequently, SP enhances robustness and mitigate the potential risks associated with weakly supervised training.

\subsection{WA inference}
\begin{figure}[hbt]
\centerline{\includegraphics[width=7cm]{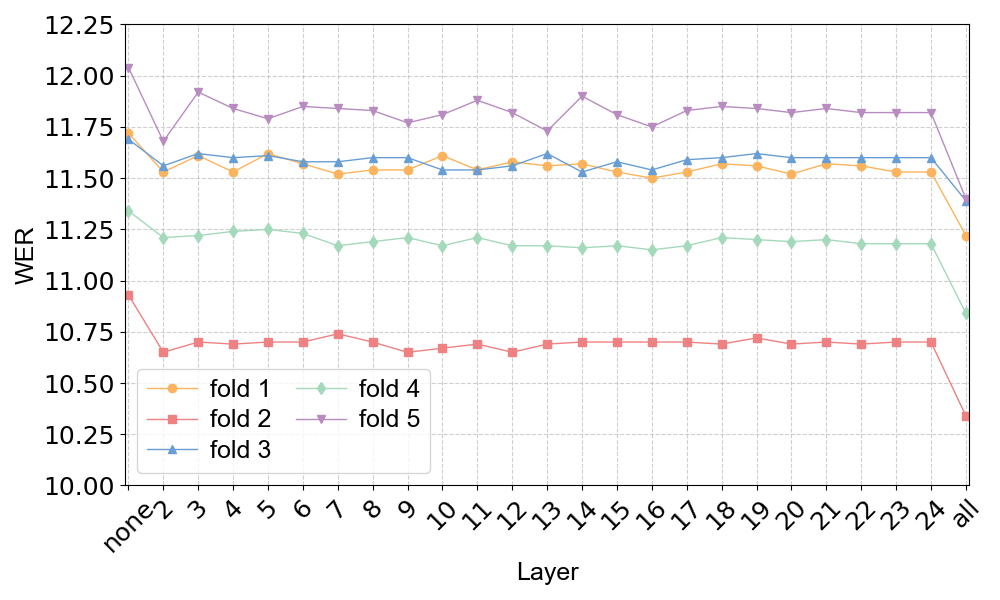}}
\caption{WERs (\%) on the 5 folds of \textit{subs-annot}, obtained by applying Gini attention weights to individual layers or all layers of transformers.}
\label{fig2}
\end{figure}

After fine-tuning with SP, the WERs of the medium and large models converge to similar levels. Therefore, subsequent experiments are conducted using the medium model for convenience. During inference, Gini coefficients are computed from the first layer's cross-attention weights, as defined in Eq.~\ref{eq1}. These coefficients are then applied to the self-attention mechanism at each or all layers to identify the most active one. The resulting WERs (\%) are presented in Fig.~\ref{fig2}.

\begin{table*}[tbh]
\centering
\caption{An example of output transcripts by using SP fine-tuning, incorporated with Gini, max or entropy-based WA inference.}
\setlength{\tabcolsep}{1.2pt} 
\renewcommand{\arraystretch}{1.2} 
\begin{tabular}{c|c}
\noalign{\hrule height 1pt}
reference & als je dan dit ziet. we kenden natuurlijk Operatie Zero. maar als je dan dit ziet hoe pijnlijk is dat dan? \\
subtitle prompt & we kenden natturlijk Operatie Zero, maar als je dit ziet, hoe pijnlijk is dat? \\
pseudo label & als je dan dit ziet, we \underline{kennen} natuurlijk \underline{operaties} zero, maar als je dan dit ziet, hoe pijnlijk is dat dan? \\
SP fine-tuning & als je dan dit ziet, we \underline{kennen} natuurlijk Operatie Zero, maar als je dan dit ziet, hoe pijnlijk is dat dan? \\
SP fine-tuning + Gini WA inference & als je dan dit ziet, we kenden natuurlijk Operatie Zero, maar als je dan dit ziet, hoe pijnlijk is dat dan? \\
SP fine-tuning + Max WA inference & als je dan dit ziet \_ \_ \_ \_ \_ \_ \_ \_ \_ \_ hoe pijnlijk is dat dan? \\
SP fine-tuning + Entropy-based WA inference & als je dan dit ziet, we \underline{kennen} \underline{de} natuurlijk Operatie Zero, maar als je dan dit ziet, hoe pijnlijk is dat dan? \\
\noalign{\hrule height 1pt}
\end{tabular}
\label{tab4}
\end{table*}

Since WA is applied solely during inference and only the test set (\textit{subs-annot}) contains verbatim transcripts, we divide this set into five folds to simulate different test scenarios and ensure robust layer selection. Each contains approximately 500 utterances, represented by different colors in the figure. Testing is performed on each fold to ensure consistent results across the dataset. 
As shown in Fig.~\ref{fig2}, WER decreases across all folds when Gini attention weights are applied, demonstrating the robustness of this mechanism across diverse subsets.
Notably, the best WER for each fold is consistently achieved when the Gini weights are applied to all layers. Under this configuration, the WER on the entire \textit{subs-annot} set improves from 11.54\% to 11.02\%.

For ablation, we also experiment with max- and entropy-based weights derived from the cross-attention weights between prompt tokens and speech frames. 
The maximum weight relies heavily on a single frame and also assigns higher weights for irrelevant tokens on average.
This makes it highly sensitive to noise and increases the risk of corruption. The results show that the max weighting leads to a high proportion of hallucinations. 
Entropy values vary significantly, with fully relevant tokens having an entropy of zero and most other tokens exhibiting large values. When normalized to $[0,1]$, many tokens with a large entropy are compressed to nearly zero, and thus lose their sensitivity. 
WER results indicate that only Gini weights effectively refine the transcript, highlighting the specificity of the association between prompt tokens and speech frames. 
Table~\ref{tab4} presents an example of output transcripts generated using Gini-, max-, and entropy-based weighted attention strategies, with errors underlined. The results show that subtitle prompts can easily refine the format and spelling of named entities in the pseudo labels.  However, errors such as ``kenden'' and ``kennen'', 
which both are syntactically and semantically correct, are more challenging to correct. 
Gini weights provide a strong cue in such cases, whereas both max- and entropy-based weights fail to refine these errors. Moreover, the max weighting tends to overlook certain speech frames, indicating its instability.

\subsection{Iterative training}
As training progresses, the model progressively improves its ability to generate accurate transcripts. The training targets are iteratively updated with newly generated transcripts at the end of each training cycle. The WERs achieved over three iterations of SP fine-tuning, followed by WA inference, are presented in Table~\ref{tab5}.

\begin{table}[hbt]
\centering
\caption{WERs (\%) across iterations after SP fine-tuning and WA inference.}
\label{table}
\setlength{\tabcolsep}{8pt} 
\renewcommand{\arraystretch}{1.2} 
\begin{tabular}{c|c|c}
\noalign{\hrule height 1pt}
 & SP fine-tuning & WA inference \\
\noalign{\hrule height 1.0pt}
iter1 & 11.54 & 11.02 \\
iter2 & 10.82 & 10.66 \\
iter3 & 10.52 & \textbf{10.34} \\
\noalign{\hrule height 1pt}
\end{tabular}
\label{tab5}
\end{table}

The WERs consistently decrease across iterations, although the rate of improvement diminishes with each subsequent iteration in both fine-tuning and inference processes. This trend suggests convergence in the iterative learning process, driven by the refinement of pseudo transcripts and subtitle prompts. To balance performance and efficiency, training is halted after three updates to the training targets. Ultimately, a WER of $10.34\%$ is achieved on the \textit{subs-annot} set, highlighting the effectiveness of the iterative approach.

\section{Conclusion}
In this paper, we propose a method to refine transcripts from existing models using subtitles, eliminating the need for any verbatim dataset. We demonstrate the effectiveness of fine-tuning models with subtitles as prompts, achieving significant improvements in WER. Analysis of rare word and OOV WERs further confirms that subtitle prompts not only enhance the model’s recall of low-frequency words but also improve generalization to OOV words. Additionally, the weighted attention mechanism we introduce further boosts transcription performance. Our results also show that the transcripts can be refined iteratively through successive training cycles. This work provides a promising direction for enhancing data quality in weakly supervised ASR systems without requiring additional labeled data.\\ 
\bibliographystyle{IEEEbib}
\bibliography{IEEEconference}

\end{document}